\documentclass[conference]{IEEEtran}

\IEEEoverridecommandlockouts
\usepackage{cite}
\usepackage{amsmath,amssymb,amsfonts}
\usepackage{graphicx}
\usepackage{textcomp}
\usepackage{xcolor}
\usepackage{multirow}
\usepackage{arydshln}
\usepackage{todonotes}
\usepackage{etoolbox}
\usepackage{setspace}
\usepackage[bottom]{footmisc}
\usepackage{placeins}
\usepackage{wrapfig}
\usepackage{color, colortbl}
\usepackage[export]{adjustbox}
\usepackage{amsmath}
\usepackage{algpseudocode}
\usepackage[ruled,vlined]{algorithm2e}
\usepackage{tikz}
\usepackage{relsize}
\usepackage{mathtools, nccmath}

\newcommand{\CC}[1]{\cellcolor{white}}
\def\BibTeX{{\rm B\kern-.05em{\sc i\kern-.025em b}\kern-0.08em
		T\kern-.1667em\lower.7ex\hbox{E}\kern-.125emX}}

\usepackage{url}

\newcommand{\fig}{Figure }
\begin{document}
	
	\title{Joint Sub-component Level Segmentation and Classification for Anomaly Detection within Dual-Energy X-Ray Security Imagery}
	
	\author{\IEEEauthorblockN{Neelanjan Bhowmik$^1$, 
			Toby P. Breckon$^{1,2}$
		}
		\IEEEauthorblockA{Department of \{Computer Science$^1$ $|$ Engineering$^2$\}, Durham University, UK}
	}
	
	\maketitle
	\begin{abstract}
		X-ray baggage security screening is in widespread use and crucial to maintaining transport security for threat/anomaly detection tasks. The automatic detection of anomaly, which is concealed within cluttered and complex electronics/electrical items, using 2D X-ray imagery is of primary interest in recent years. We address this task by introducing joint object sub-component level segmentation and classification strategy using deep Convolution Neural Network architecture. The performance is evaluated over a dataset of cluttered X-ray baggage security imagery, consisting of consumer electrical and electronics items using variants of dual-energy X-ray imagery (pseudo-colour, high, low, and effective-Z). The proposed joint sub-component level segmentation and classification approach achieve $\sim99\%$ true positive and  $\sim5\%$ false positive for anomaly detection task.
	\end{abstract}
	

	\section{Introduction} \label{s:intro}
	
\begin{figure}
		\centering
		\includegraphics[width=\linewidth, cfbox=gray]{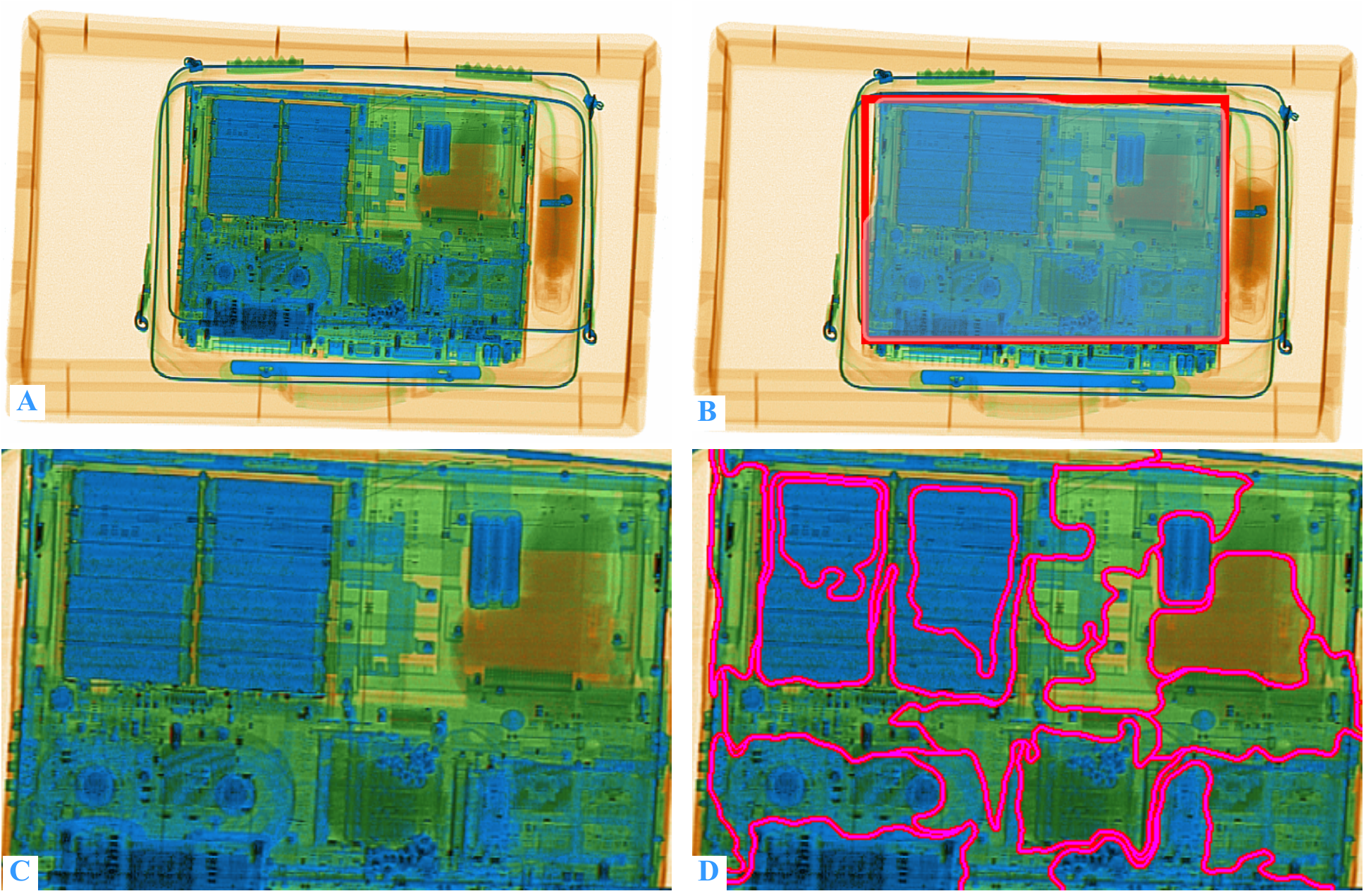}
		\caption{Exemplar 2D X-ray imagery (A) used for object level anomaly detection (B/C) via mask R-CNN segmentation and sub-component level anomaly detection (D) via joint object over-segmentation and classification.}
		\label{fig:block_algo}
	\end{figure}

	With the increasing volume of traffic, we need to ensure an efficient system for aviation securing capable of addressing the evolving threat landscape that emanates from broader global geopolitical events. Currently multiple-view X-ray baggage security screening is widely used to maintain aviation and transport including the screening of electronics/electrical items. To address the future challenges of increasing volumes and complexities, the recent focus on the use of automated screening approaches is of particular interest. This includes the potential for automatic anomaly/threat detection as a methodology for concealment detection within complex electronics and electrical items screened using low-cost, 2D X-ray imagery. Passenger baggage is currently inspected manually using dual-energy multiple-view X-ray imaging. The threat concealment can also be very subtle and very well hidden (Figure \ref{fig:block_algo}A) challenging for a human operator to identify.
	
	Early work on automated threat detection within X-ray security images is based on hand-crafted features \cite{turcsany2013xray,Bastan2013ObjectRI} (Bag-of-Visual-Words), which is applied together with a classifier such as a Support Vector Machine. 
	More recent work \cite{Akcay2017:XrayICIP, Akcay2018:XrayIEEETransaction}, that specifically leverage recent advances in Convolutional Neural Networks (CNN) deep learning architectures  \cite{alexnet,He2015:ResNet}, have now been shown to outperform earlier approaches in terms of true positive detection, false alarm rate and the range of objects that can be detected in a side by side comparison. 
	By contrast, CNN approach of \cite{Akcay2017:XrayICIP} operates at scan rate ($<$1 sec. per image), with higher accuracy and targets probabilistic item localisation within each X-ray view.  The work of \cite{griffin2018unexpected} considers a unique feature representation as a critical component for detection within cluttered X-ray imagery for anomaly detection. In the works of \cite{Akcay2018:GANomaly,Akcay2019:skipganomaly}, semi-supervised anomaly detection strategies are proposed based on high reconstruction errors produced by a generator network adversarially trained on benign X-ray imagery only.
	
	\begin{figure*}[!htb]
		\centering
		\includegraphics[width=\linewidth,  cfbox=gray]{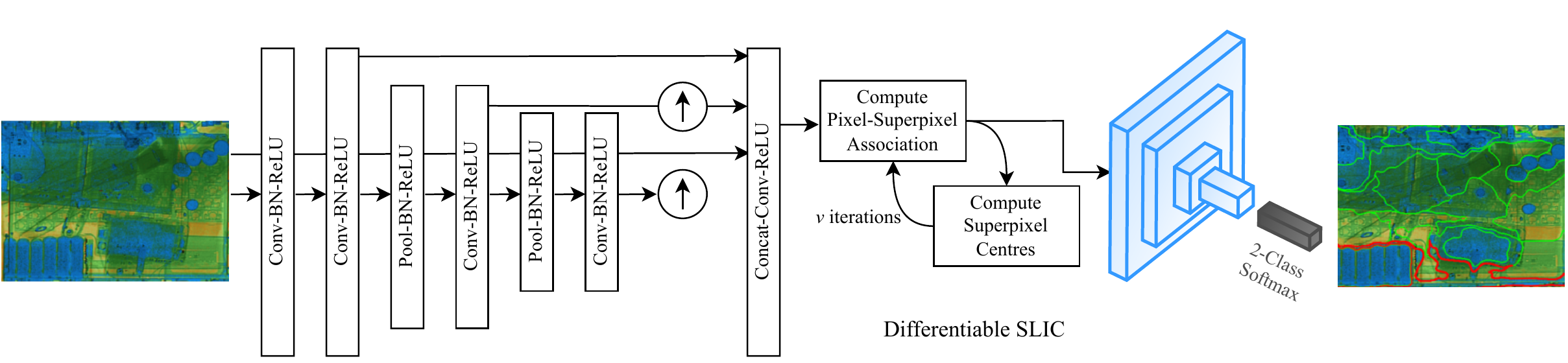}
		\caption{End-to-end joint segmentation and classification CNN architecture. Each segment (green/red contours) is extracted using differentiable SLIC prior to classification.}
		\label{fig:diff_slic}
	\end{figure*}
	
	Earlier superpixel algorithms can be classified into clustering and graph based strategies. Clustering strategies \cite{Achanta2012:SLIC, li15:superpixel}, leverage traditional clustering techniques such as k-means for superpixel segmentation. Graph based approaches \cite{ren03:NC, felzenszwalb04:seg} define the superpixel over-segmentation as a graph partitioning problem, in which nodes are represented by pixels and the edges denote the strength of connectivity between adjacent pixels. 
	Most of these methods rely on traditional hand-crafted features and do not use deep CNN techniques. More recent approach, such as in \cite{tu18:afinityloss}, deep features are used for superpixel segmentation bypassing the gradients through non-differentiable superpixel algorithms. CNN based unsupervised data clustering approaches are proposed \cite{rasmus2015:semi, xie16:unsupervised} in literature. 
	A deep embedded clustering framework, for simultaneously learning feature representations and cluster assignments is proposed in the work of \cite{xie16:unsupervised}. A more recently, an end-to-end deep learning-based clustering algorithm called Superpixel Sampling Networks (SSN) \cite{jampani18:superpixel} is developed for superpixel segmentation task where it can use image-specific constraints. However, none of the above discussed works is targeted for X-ray image application specific tasks. Following the work in \cite{Zhang2014:XrayCargo}, in our work we leverage the use of superpixels \cite{Achanta2012:SLIC, jampani18:superpixel} within X-ray security baggage image classification with prior object segmentation \cite{He2017:MaskRCNN} as an enabler to CNN based end-to-end joint sub-component level anomaly detection within X-ray security imagery.
	
	In this work, we evaluate two automatic segmentation strategies for intra-object anomaly detection in X-ray security imagery, as illustrated in the \fig \ref{fig:block_algo}A:- 
	\begin{itemize}
		
		\item [--] First, object level segmentation is performed (\fig \ref{fig:block_algo}B $\rightarrow$ \fig \ref{fig:block_algo}C).
		
		\item [--] Secondly, end-to-end joint segmentation and classification CNN architecture is proposed (\fig \ref{fig:block_algo}C $\rightarrow$ \fig \ref{fig:block_algo}D) for anomaly detection as a binary, $\{anomaly, benign\}$, classification task.
		
		\item [--] Additionally, 
		we study the impact of dual-energy X-ray imagery (pseudo-colour, high, low and effective-Z) for anomaly detection task.
		
	\end{itemize}

	\section{Proposed Approach} \label{s:proposal}
	
	We outline the approach of this paper in the following section: object level localisation using Mask R-CNN \cite{He2017:MaskRCNN} in Section \ref{ss:obj}, followed by CNN based end-to-end segmentation and classification strategy in Section \ref{ss:e2e}.
	
	\subsection{Object Level Localisation} \label{ss:obj}
	We consider contemporary CNN architectures, such as Mask R-CNN \cite{He2017:MaskRCNN}, Faster R-CNN \cite{Ren2015:FasterRCNN}, for object detection and segmentation task to explore their applicability for generalised object detection/instance segmentation tasks within the context of X-ray security baggage imagery. 
	Mask R-CNN \cite{He2017:MaskRCNN} relies on region proposals followed by ROI-Pooling to produce standard-sized outputs, which include pixel-wise image mask of a detected object, suitable for input into a secondary classifier. It addresses feature map misalignment of Faster R-CNN \cite{Ren2015:FasterRCNN} by incorporating bilinear boundary interpolation. Mask R-CNN combines object localisation with instance segmentation of the object in the image (\fig \ref{fig:block_algo}B $\rightarrow{}$ \fig \ref{fig:block_algo}C). This architecture \cite{He2017:MaskRCNN} is evaluated over an electronics and electrical items packed within cluttered X-ray security baggage, for anomaly detection task.
	
	\subsection{Joint Segmentation and Classification via CNN} \label{ss:e2e}
	For object over-segmentation (\fig \ref{fig:block_algo}D), we apply deep CNN based object sub-component level segmentation method, Super Sampling Network (SSN) \cite{jampani18:superpixel} combined with classification network for $\{anomaly, benign\}$ classification. At the core of SSN is a differentiable clustering technique, which is inspired by Simple Linear Iterative Clustering (SLIC) \cite{Achanta2012:SLIC} approach. SLIC performs iterative clustering, where initially image is segmented into roughly equal sized segments. To measure the similarity between the segments, it introduces a new distance metric which considers the size of the segment. It takes the user define number of approximately equally-sized superpixel $K$. For an image with $N$ pixels, the approximate size of each superpixel will be $N/K$. 
	
	The core of part of SLIC \cite{Achanta2012:SLIC} is iterative clustering, which is non-differentiable due to the computation of pixel to superpixel associations involving nearest neighbour operation. Instead of computing hard pixel-superpixel associations, SSN approach computes soft-associations between pixels and superpixels, which makes it differentiable.
	
	\fig \ref{fig:diff_slic} depicts the joint segmentation and classification architecture. In SSN, the CNN for feature extraction is composed of a series of convolution layers interleaved with batch normalisation and ReLU activation. The max-pooling layer down samples the input by factor 2, after 2$^{nd}$ and 4$^{th}$ convolutional layer output. Convolution filter size of $3\times3$ is used with the number of output channels set to 64 in each layer. Other CNN architectures can also be integrated easily in this framework. The final features are passed onto the differentiable SLIC module, which iteratively updates the pixel-superpixel association and computes superpixel centres.  We use reconstruction loss ($l_{rcon}$), which is cross-entropy loss, and compactness loss ($l_{comp}$) to reduce the spacial variance in each superpixel cluster. The total loss is the sum of the above two loss functions $(L = l_{rcon} + \lambda l_{comp}, where \hspace{0.2cm} \lambda = 1e-4)$. 
	
	Followed by this stage, the superpixels are fed into a binary classifier (as shown in the \fig \ref{fig:diff_slic}) for anomaly detection in each segment. Each segmented image region, sub-component level segmentation, is subsequently classified using a deep CNN architecture model formulated as a binary, $\{anomaly, benign\}$, classification task.
	
	\noindent \textit{SqueezeNet} \cite{Iandola2016:SqueezeNet} is a small network architecture that uses many 1-by-1 filters to aggressively reduce the number of weights. It offers equivalent accuracy to the AlexNet \cite{alexnet} yet operating with $50\times$ fewer parameters. 
	
	\noindent \textit{VGG} \cite{Simonyan2014:VGGverydeep} is a seminal network architecture that consists of several deep convolutional layers (e.g. $11-19$ weight layers), with a fixed kernel size $(3 \times 3)$ (convolution stride is set to 1), stacked on top of each other in increasing depth, which shows notable performance improvements on prior architecture \cite{alexnet,ZFNet}.
	
	\noindent \textit{ResNet} \cite{He2015:ResNet} solves the issue of vanishing gradient present in the forward feed and backward propagation processing in previous CNN architectures by introducing skip connection, parallel to the regular convolutional layers ($18-152$ in depth).
	
	\section{Evaluation} \label{s:eval}
	This section presents the dataset used, the implementation details, and the results of our experiments.
	
	\subsection{Experimental Setup} \label{ssub:exp_setup}
	Our experimental setup comprises of following dataset.
	\begin{figure}[tb]
		\centering
		\includegraphics[width=\linewidth]{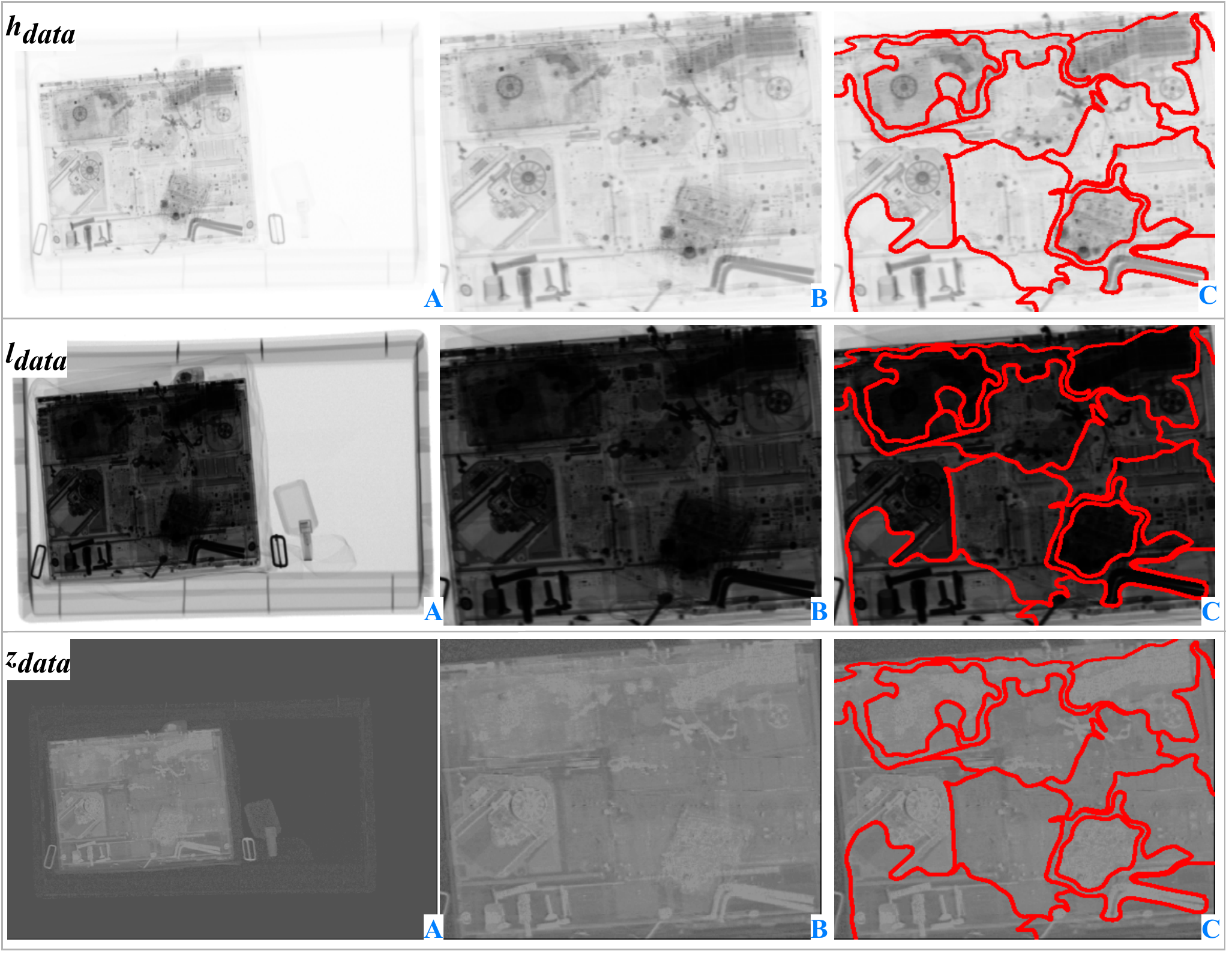}
		\caption{Exemplar high, low, and effective-Z X-ray imagery from {\it DEEi} dataset (A) used for object level (B) and sub-component level segmentation (C) $\{anomaly, benign\}$ classification.}
		\label{fig:ex_hlz}
	\end{figure}
	
	\noindent {\it \textbf{DEEi. }} The dataset ({Durham Electrical and Electronics Items}) is constructed using a 2D X-ray scanner with associated pseudo-colour materials mapping via dual-energy. All X-ray imagery is gathered locally by using a dual-energy X-ray scanner (Gilardoni FEP ME 640 AMX) \cite{gilardoni}. The dataset consists of large consumer electronics (e.g. laptop, mobile) and electrical (e.g. iron, hairdryer, toaster) items with and without anomaly (e.g. marzipan, screws, metal plates, sharps, etc.) concealment present as illustrated in Figures \ref{fig:block_algo}A. Our dataset consists of disparate models and shapes of electronics items, where the inserted anomaly concealment might be challenging to identify. In total, we use $7,022$ X-ray imagery ($70:30$ data split) for our experiment. Additionally, we access the dual-energy X-ray (`raw'), data (three types - high ($h_{data}$), low ($l_{data}$) energy, and effective-Z ($z_{data}$) response (\fig \ref{fig:ex_hlz})) from the X-ray scanner \cite{gilardoni} for $\{anomaly, benign\}$ classification.
	
	Training for all architecture variants is performed via transfer learning using stochastic gradient descent with a momentum of $0.9$, a learning rate of $0.0002$, a batch size of $64$. All networks are trained on NVIDIA $1080$Ti GPU via the PyTorch framework \cite{pytorch}.

	\subsection{Evaluation of Object and Sub-component Level Classification} \label{evaluation}
	Our model performances are evaluated in terms of Accuracy (A), Precision (P), F-score (F1), True Positive (TP\%), and False Positive (FP\%), as presented in the Tables  \ref{Table:object} and \ref{Table:sp} where we additionally compare our `raw' X-ray data approach to the use of the pseudo-colour imagery conventionally used in automated object detection studies in X-ray images \cite{gaus2019evaluation,Akcay2017:XrayICIP,Akcay2018:GANomaly}.
	
	\begin{table}[ht]
\centering
\caption{Object level segment and classification using varying CNN architectures with pseudo-colour and `raw' X-ray data.}
\addtolength{\tabcolsep}{-3.4pt}
\begin{tabular}{llllllll}
\hline
\rowcolor{blue!10}
& Data & Network & A & P & F1 & TP(\%) & FP(\%)\\ \hline 
\multirow{12}{*}{\rotatebox[origin=c]{90}{\shortstack[l]{Object level\\ segmentation}}} & \multirow{3}{*}{\shortstack[l]{\it pseudo- \\ \it colour}} 
& \CC{40}\small SqueezeNet &\CC{40}0.83 & \CC{40}0.78 & \CC{40}0.82 & \CC{40}93.14 & \CC{40}26.97 \\
&& \CC{20}\small VGG-16  & \CC{20}0.76  & \CC{20}0.69 & \CC{20}0.75 & \CC{20}94.26 & \CC{20}39.47 \\ 
&& \CC{40}\small ResNet$_{50}$ &\CC{40}{\bf 0.86} & \CC{40}{\bf 0.84} & \CC{40}{\bf 0.84} & \CC{40}\bf{97.29} & \CC{40}{\bf 16.59} \\
\cline{2-8}

& \multirow{3}{*}{$h_{data}$} & \CC{40}\small SqueezeNet &\CC{40}0.82 & \CC{40}0.76  &\CC{40}0.81 & \CC{40}92.67 & \CC{40}28.01 \\
&& \CC{20}\small VGG-16  & \CC{20}0.75  & \CC{20}0.67  & \CC{20}0.73 & \CC{20}96.76  & \CC{20}45.46 \\ 
&& \CC{40}\small ResNet$_{50}$  &\CC{40}0.85 & \CC{40}0.80 &\CC{40}0.85 & \CC{40}94.83 & \CC{40}22.24 \\
\cline{2-8}

& \multirow{3}{*}{$l_{data}$} & \CC{40}\small SqueezeNet &\CC{40}0.75 & \CC{40}0.69  &\CC{40}0.75 & \CC{40}88.64 & \CC{40}35.28 \\
&& \CC{20}\small VGG-16  & \CC{20}0.71 & \CC{20}0.63  & \CC{20}0.68 & \CC{20}97.18  & \CC{20}55.01 \\ 
&& \CC{40}\small ResNet$_{50}$  &\CC{40}0.81 & \CC{40}0.80 &\CC{40}0.64  & \CC{40}84.84 & \CC{40}20.76 \\
\cline{2-8}

&\multirow{3}{*}{$z_{data}$} & \CC{40}\small SqueezeNet &\CC{40}0.76 & \CC{40}0.73  &\CC{40}0.65 & \CC{40}80.31 & \CC{40}21.32 \\
&& \CC{20}\small VGG-16  & \CC{20}0.68  & \CC{20}0.66  & \CC{20}0.64 & \CC{20}84.58 & \CC{20}47.73 \\ 
&& \CC{40}\small ResNet$_{50}$  &\CC{40}0.84 & \CC{40}0.80 &\CC{40}0.81  & \CC{40}90.62 & \CC{40}20.45 \\
\hline

\end{tabular}
\label{Table:object}
\addtolength{\tabcolsep}{-0pt}
\end{table}
	\begin{table}[ht]
\centering
\caption{Object sub-component level joint segmentation and classification with pseudo-colour and `raw' X-ray data.}
\addtolength{\tabcolsep}{-3.4pt}
\begin{tabular}{llllllll}
\hline
\rowcolor{blue!10}
 & Data & Network & A & P & F1 & TP(\%) & FP(\%)\\ \hline
\multirow{15}{*}{\rotatebox[origin=c]{90}{\shortstack[l]{Sub-component \\ level segmentation}}} & \multirow{3}{*}{\shortstack[l]{\it pseudo- \\ \it colour}} 
& \CC{40}\small SqueezeNet &\CC{40}0.95 &\CC{40}0.92 &\CC{40}0.94 &\CC{40}99.10  &\CC{40}8.90 \\
&& \CC{20}\small VGG-16  & \CC{20}0.93 &\CC{20}0.91  &\CC{20}0.93 &\CC{20}95.89  &\CC{20}8.55 \\ 
&& \CC{40}ResNet$_{50}$  &\CC{40}{\bf 0.97} &\CC{40}{\bf 0.95} &\CC{40}{\bf 0.97} &\CC{40}98.99 &\CC{40}{\bf 4.54} \\
\cline{2-8}

& \multirow{3}{*}{$h_{data}$} & \CC{40}\small SqueezeNet &\CC{40}0.96 & \CC{40}0.94  &\CC{40}0.96 & \CC{40}98.86 & \CC{40}6.12 \\
&& \CC{20}\small VGG-16  & \CC{20}0.96  & \CC{20}0.94 & \CC{20}0.96 & \CC{20}98.71  & \CC{20}6.01 \\ && \CC{40}ResNet$_{50}$  &\CC{40}0.96 & \CC{40}0.94 &\CC{40}0.96  & \CC{40}99.79 & \CC{40}6.16 \\
\cline{2-8}

&\multirow{3}{*}{$l_{data}$} & \CC{40}\small SqueezeNet &\CC{40}0.93 & \CC{40}0.91  &\CC{40}0.93 & \CC{40}96.85 & \CC{40}9.53 \\
&&\CC{20}\small VGG-16  & \CC{20}0.95  & \CC{20}0.93  & \CC{20}0.98 & \CC{20}98.76  & \CC{20}7.09 \\ 
&&\CC{40}ResNet$_{50}$  &\CC{40}0.96 & \CC{40}0.93 &\CC{40}0.95  & \CC{40}98.64 & \CC{40}6.37 \\
\cline{2-8}

&\multirow{3}{*}{$z_{data}$} & \CC{40}\small SqueezeNet &\CC{40}0.90 & \CC{40}0.87  &\CC{40}0.89 & \CC{40}95.28 & \CC{40}15.06 \\
&&\CC{20}\small VGG-16  & \CC{20}0.95  & \CC{20}0.93  & \CC{20}0.95 & \CC{20}97.18  & \CC{20}6.52 \\ 
&&\CC{40}ResNet$_{50}$  &\CC{40}0.96 & \CC{40}0.94 &\CC{40}0.96  & \CC{40}98.99 & \CC{40}5.93 \\
\cline{2-8}

&\multirow{3}{*}{$hlz_{data}$} & \CC{40}\small SqueezeNet &\CC{40}0.96 &\CC{40}0.94 &\CC{40}0.96 &\CC{40}99.74 &\CC{40}6.26 \\
&&\CC{20}\small VGG-16  &\CC{20}0.96 &\CC{20}0.94 &\CC{20}0.96 &\CC{20}99.75 &\CC{20}6.03 \\ 
&&\CC{40}ResNet$_{50}$  &\CC{40}{\bf 0.97}  &\CC{40}0.94  &\CC{40}0.94  &\CC{40}{\bf 100}  &\CC{40}6.07 \\
\hline

\end{tabular}
\label{Table:sp}
\addtolength{\tabcolsep}{-0pt}
\vspace{-0.3cm}
\end{table}

	In the object-level segmentation strategy (Table \ref{Table:object}), where the target object is first detected, localised and isolated via segmentation prior to binary classification, the maximum accuracy is achieved with ResNet$_{50}$ (A: $0.86$, TP: $97.29\%$ - Table \ref{Table:object}, {\it pseudo-colour}), due to the focused feature representation. We observe that the use of `raw' X-ray data achieves good true positive (TP: $97.18\%$ - Table \ref{Table:object}, $l_{data}$), but suffers with relatively high false positive rate. The lowest FP is $16.59\%$ (with ResNet$_{50}$ - Table \ref{Table:object}, {\it pseudo-colour}), but fails to outperform object sub-component level strategy (FP: $4.54\%$ - Table \ref{Table:sp}, {\it pseudo-colour}).
	
	From the results presented in Tables \ref{Table:object} and \ref{Table:sp}, we observe from the two strategies considered that the joint sub-component level segmentation and classification strategy for anomaly detection via ResNet$_{50}$, offers significantly superior anomaly detection performance (A: $0.97$, TP: $98.99\%$, FP: $4.54\%$ - Table \ref{Table:sp}, {\it pseudo-colour}) compared to an object level segmentation strategy overall (Table \ref{Table:object}). Although SqueezeNet achieves the maximum true positive (TP: $99.10\%$), but it suffers relatively high false positive (FP: $8.90\%$) when using pseudo-colour X-ray imagery as presented in Table \ref{Table:sp} - {\it pseudo-colour}. When we use `raw' X-ray data for classification, we achieve the highest true positive (TP: $100\%$, A: $0.97$, Table \ref{Table:sp}, $hlz_{data}$) with ResNet$_{50}$ and combination of high, low and effective-Z energy data, but this is effected by the relatively high false positive rate (FP: $6.07\%$, Table \ref{Table:sp}, $hlz_{data}$). Overall, ResNet$_{50}$ performs the best across all the three (high, low, effective-Z and combination) `raw' X-xay data with TP $> 99\%$.  Overall the sub-component level segmentation provides higher granularity information compared to the object level segmentation, henceforth achieves the best performance.
	
	To our knowledge, the proposed work on joint end-to-end object sub-component level segmentation and classification is one of the first works for the anomaly detection task. Therefore, we consider two segmentation strategies, i.e., object-level (using Mask R-CNN)vs sub-component level to compare the results (Tables \ref{Table:object}, \ref{Table:sp}).

\begin{figure}[!htb]
		\centering
		\includegraphics[width=\linewidth]{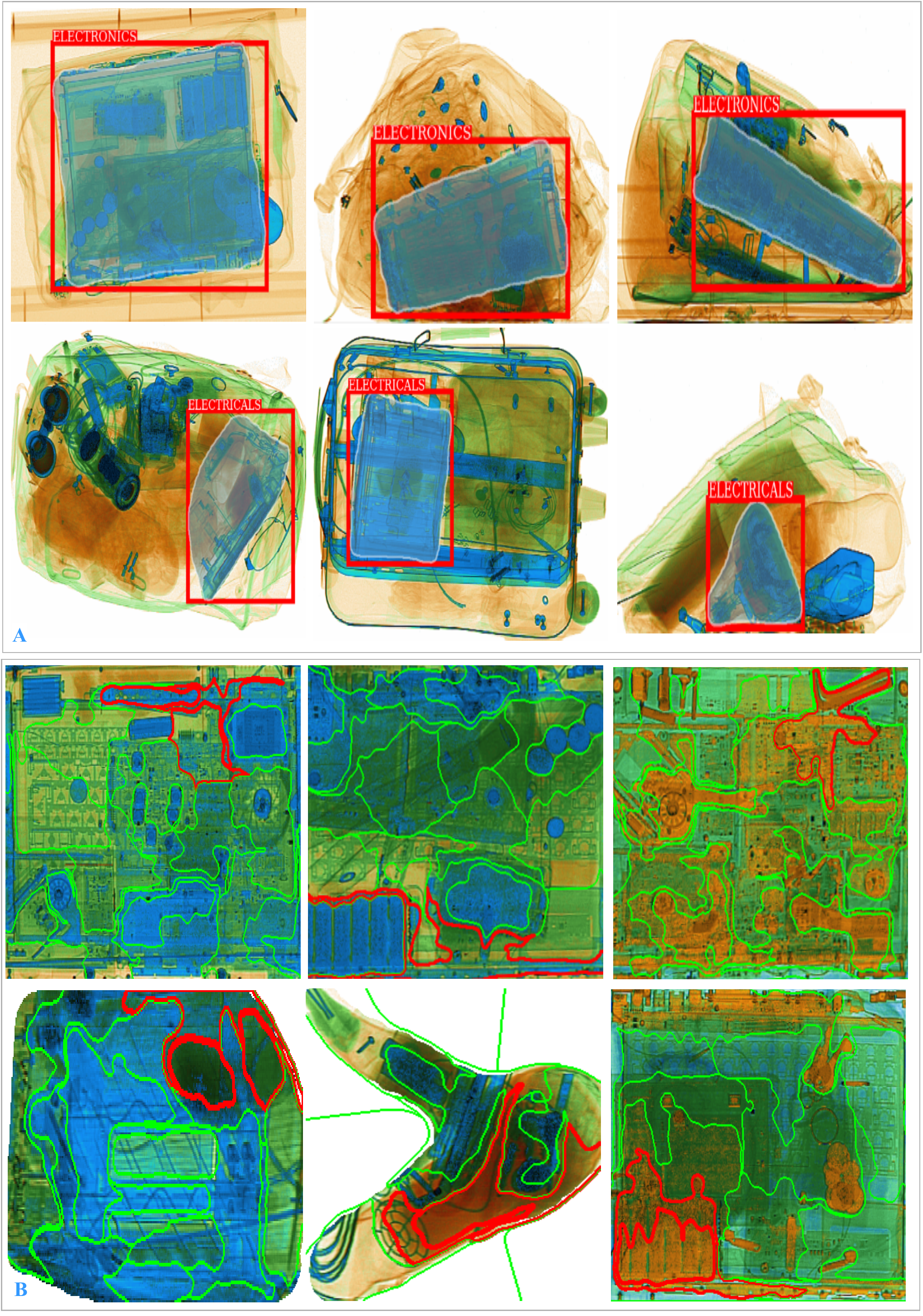}
		\vspace{-0.8cm}
		\caption{Examples of $\{anomaly, benign\}$ detection and classification from {\it DEEi} dataset: (A) object level segmentation and (B) joint object sub-component segmentation (contours) and  classification (\textcolor{green}{green}: benign, \textcolor{red}{red}: anomaly).}
		\label{fig:result_example}
	\end{figure}

\begin{figure}[!htb]
		\centering
		\includegraphics[width=\linewidth]{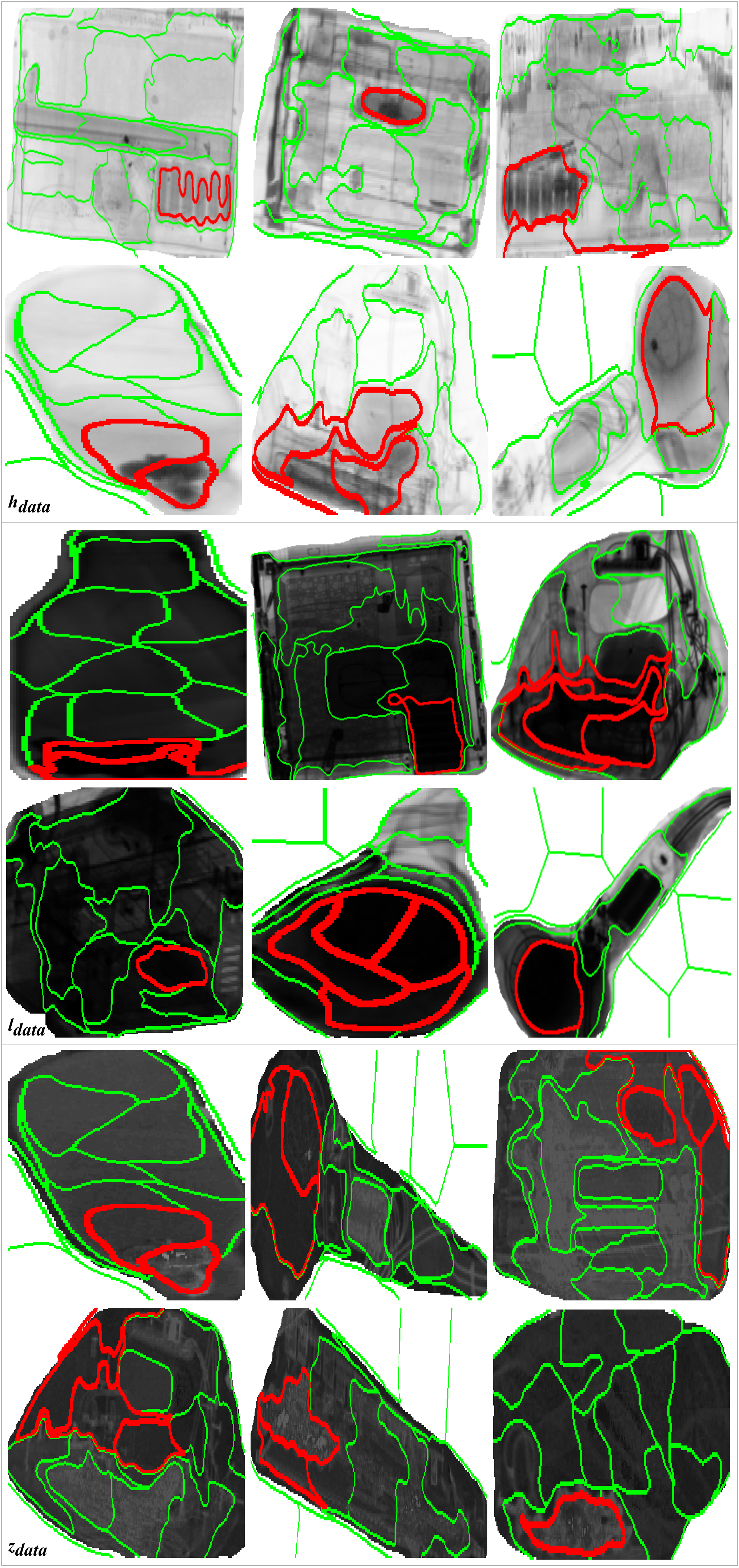}
		\vspace{-0.8cm}
		\caption{Examples anomaly detection using high, low, and effective-Z X-ray imagery (`raw') from {\it DEEi} dataset: object sub-component segmentation (contours) and classification (\textcolor{green}{green}: benign, \textcolor{red}{red}: anomaly).}
		\label{fig:raw_result_example}
	\end{figure}

	Examples of the detection (object level segmentation via Mask R-CNN) and classification of the consumer electrical and electronics items containing an anomaly are depicted in \fig \ref{fig:result_example}A. \fig \ref{fig:result_example}B illustrates exemplary qualitative results of joint sub-component segmentation and classification of electrical and electronics items, where red colour indicates anomalous region while green represents benign sub-components. 
	Examples of anomaly detection using `raw' energy X-ray data are depicted in the \fig \ref{fig:raw_result_example}.
	The benefit of using sub-component level segmentation is it's capability to provide precise localisation of the anomalous region within complex object (Figures \ref{fig:result_example}B, \ref{fig:raw_result_example}).
	

	\section{Conclusion} \label{s:conclusion}
	
	We evaluate the performance impact of two different strategies, object segmentation, and object sub-component segmentation, for concealed threat/anomaly detection within consumer electrical/electronics item using deep CNN based end-to-end architecture. Our experimental results exhibit that the best performance ($>99\%$ TP and $\sim5\%$ FP) is achieved with object sub-component segmentation strategy on CNN classifier using `raw' X-ray data (a combination of high, low energy and effective-Z). To the best of our knowledge, the use of `raw' X-ray imagery for anomaly detection tasks is one of the first and novel of its kind. Therefore, our study possibly opens up a plethora of broad research interests in the X-ray imagery domain. Within the context of electrical and electronics items, this work offers the automatic first-stage screening of aviation baggage for anomalous item detection at the component level as an indicator of potential threat presence. 
	In our future work on anomaly detection, we primarily focus on combining multiple data (a combination of pseudo-colour and high, low energy and effective-Z) to achieve higher accuracy and lower false positive. Additionally, we will target varied electronic and electrical items across a full range of operational X-ray characteristics.

	
	\begingroup
	\setstretch{.96}
	
	\bibliographystyle{IEEEtran}
	\bibliography{egbib,object_detection,ref_mod,ref_1,IEEEabrv,IEEEreference}
	
	\endgroup

\end{document}